\documentclass[letterpaper, 10 pt, conference]{ieeeconf}

\IEEEoverridecommandlockouts
\usepackage{graphicx}
\usepackage{capt-of} 
\usepackage{amsmath}
\usepackage{amssymb}
\usepackage{cite}
\usepackage{booktabs}
\usepackage{multirow}
\usepackage{array}
\usepackage{algorithm}
\usepackage{algorithmic}
\usepackage[table]{xcolor}
\usepackage{hyperref}

\makeatletter
\renewcommand\fs@ruled{\def\@fs@cfont{\bfseries}\let\@fs@capt\floatc@ruled
  \def\@fs@pre{\kern8pt\hrule height.8pt depth0pt \kern2pt}%
  \def\@fs@post{\kern2pt\hrule\relax}%
  \def\@fs@mid{\kern2pt\hrule\kern2pt}%
  \let\@fs@iftopcapt\iftrue}
\makeatother

\title{\LARGE \bf
Air-Ground Collaborative Vision-and-Language \\
Navigation via Shared Bird's-Eye Maps
}

\author{Shuning Zhang, Liang Li, Yunheng Wang, Tao Wang, Yihang Kang, Renjing Xu%
\thanks{Shuning Zhang and Liang Li have the same contributions.}
\thanks{Shuning Zhang, Yunheng Wang, Tao Wang, Yihang Kang and Renjing Xu are with the Robotics and Autonomous Systems Thrust, Systems Hub, The Hong Kong University of Science and Technology (Guangzhou), China
{\tt\small szhang272@connect.hkust-gz.edu.cn, yunhengwang1214@gmail.com, 19819824568@163.com, KYH2107405729@163.com, renjingxu@hkust-gz.edu.cn}}%
\thanks{Liang Li is with the School of Communications and Information Engineering, Nanjing University of Posts and Telecommunications, China
{\tt\small lianglee2004@163.com}}%
}

\hypersetup{
  pdfauthor={Shuning Zhang},
  pdftitle={Air-Ground Collaborative VLN via Shared Bird's-Eye Maps},
  pdfsubject={arXiv preprint},
  pdfkeywords={}
}

\begin{document}

\IEEEaftertitletext{%
  \begin{minipage}{\textwidth}
    \centering
    \includegraphics[width=0.99\textwidth]{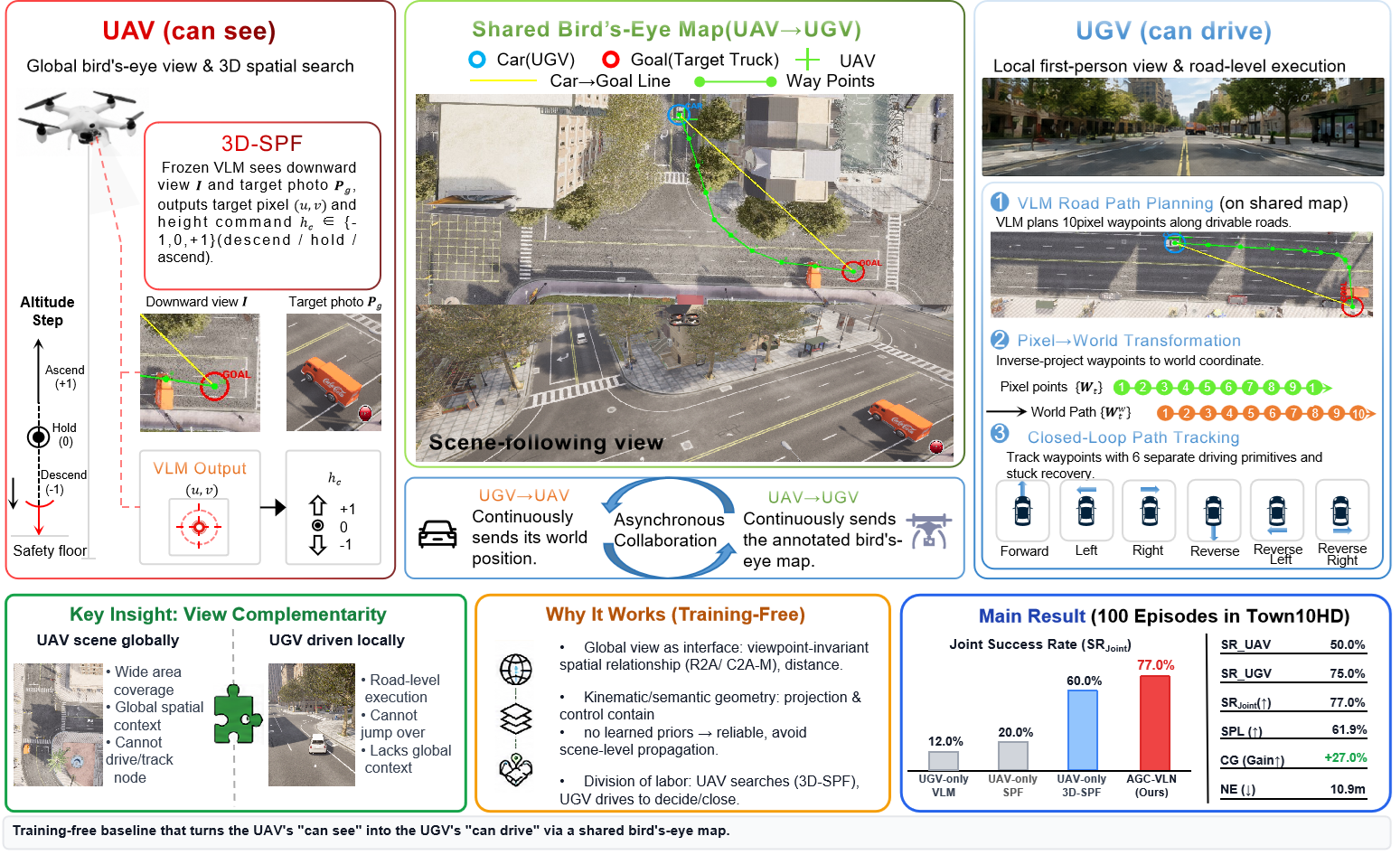}
    \captionof{figure}{The UAV turns its ``can see'' into the UGV's ``can
    drive''. View asymmetry: the UAV sees the whole block from a global
    bird's-eye view while the UGV drives with a local first-person view.
    Shared map: the UAV renders the UGV pose and the target's position
    as \texttt{CAR}/\texttt{GOAL} markers into a shared bird's-eye map.
    Parallel execution: the UAV flies 3D-SPF toward the target while the
    UGV follows a VLM road path on the map, and either arriving counts as
    success, a training-free air-ground collaboration.}
    \label{fig:teaser}
     \vspace{0.3cm}
  \end{minipage}%
}

\maketitle
\thispagestyle{empty}
\pagestyle{empty}

\begin{abstract}

Air-ground collaborative Vision-and-Language Navigation (VLN) pairs an
unmanned aerial vehicle (UAV) with a global bird's-eye view and an unmanned
ground vehicle (UGV) with a local first-person view, yet the setting remains
largely unexplored: existing training-free methods solve single-agent tasks
but offer no collaboration mechanism, and a recent CARLA-Air evaluation found
no stable cooperative behavior across five state-of-the-art VLA models; naive
semantic communication or bidirectional coupling even \textit{degrades}
performance.
We establish \textbf{AGC-VLN} (Air-Ground Collaborative VLN), the first
training-free baseline for air-ground collaborative VLN. The key insight is
that training-free methods decompose navigation into VLM-based semantic
reasoning and deterministic geometric execution, exposing a collaboration
interface: the UAV's global view, over which it renders the UGV's reported
pose and the target's position as \texttt{CAR}/\texttt{GOAL} markers with
distance labels, yielding a \textbf{shared bird's-eye map}. From this map, the
UGV acquires global spatial context its first-person view cannot provide,
plans a road-following path with a frozen VLM, and executes it under
closed-loop control; in parallel, the UAV runs \textbf{3D-SPF}, a
spatial-search upgrade of SPF that localizes the target in the downward view
and flies toward it.
On 100 closed-loop episodes in CARLA-Air's Town10HD scene, AGC-VLN reaches a
\textbf{77.0\%} joint success rate, a collaboration gain of \textbf{+27.0\%}
over the weaker individual agent (the UAV, 50.0\%), and exceeds the strongest
published single-agent baseline (Travel UAV, 53.0\%) by 24.0 points, stemming
from the complementarity of the UAV's global view and the UGV's road-following
execution. Project page: \url{https://github.com/ZSN2024/AGC-VLN}.

\end{abstract}

\section{INTRODUCTION}

Frozen vision-language models have rewritten the recipe for
Vision-and-Language Navigation (VLN).
The classic benchmarks (R2R~\cite{r2r}, REVERIE~\cite{reverie}) belonged to
an era of training a dedicated policy per dataset. See-Point-Fly
(SPF)~\cite{spf2025} already flies a real drone to 92.7\% success by merely
asking a frozen VLM to ``point'' at the image, with no training at all.
Uni-LaViRA~\cite{unilavira}, Fly0~\cite{fly0}, and
FineCog-Nav~\cite{finecognav} carry the same recipe across embodiments and
reasoning styles.
One assumption, however, survives intact from the classic era:
\textit{a single robot navigates alone}.

Multi-robot collaborative navigation is the natural next step, and
heterogeneous \textbf{air-ground teams} are its most representative form.
Search-and-rescue, last-mile delivery, and escort need wide-area
coverage from the air plus ground-level reachability from a wheeled platform;
pairing the two promises a division of labor no single embodiment can
replicate.
The infrastructure has recently arrived: CARLA-Air~\cite{carlaair_infra}
unifies CARLA~\cite{carla} and AirSim~\cite{airsim} in a single Unreal
Engine process with zero-latency sensor synchronization, and
AirGroundBench~\cite{airgroundbench} provides 115 closed-loop air-ground
VLN episodes across 11 environments.

The CARLA-Air cooperation study~\cite{carlaair_vla} evaluated five aerial VLA
models (AerialVLA, OpenFly, OpenUAV, SPF, AerialVLN) on closed-loop
air-ground tasks, and the verdict was blunt: \textbf{none of them could turn
single-agent skill into cooperative behavior}.
Worse, naive communication \textit{hurt}: UGV text hints \textit{degraded}
most models, bidirectional coupling amplified errors for all, and even oracle
geometric cues did not close the gap.
The study attributes this to missing partner-state anchoring,
low-latency action coordination, and team-level objective alignment, while a
rule-based controller showed the tasks are solvable.

In short: \textit{the platform exists, the benchmark exists, the failure is
diagnosed, but no working air-ground collaborative VLN method has been
demonstrated.}

In this paper we establish \textbf{AGC-VLN} (Air-Ground Collaborative VLN),
deliberately a \textbf{training-free baseline}.
Our key insight is that the very decomposition that makes training-free
methods work for a single agent also solves the collaboration problem.
SPF-style pipelines separate \textit{semantic reasoning} (a frozen VLM marks
the target in the image) from \textit{geometric execution} (deterministic
projection and control).
We keep this decomposition but extend the VLM's action interface. Beyond the
2D point, the VLM also outputs a discrete height command (descend, hold, or
ascend). This gives the UAV active search in the vertical dimension and
upgrades SPF's fixed-altitude planar point-fly into
\textbf{three-dimensional spatial search}.
This spatial-aware, 3D-action-interface refinement of SPF is our
\textbf{3D-SPF}.
The seam between the two stages is a metric quantity: the
\textit{viewpoint-invariant} \textbf{spatial annotation} in the UAV's
bird's-eye view, where the UAV renders the teammate's reported pose and the
target's position into a shared map the UGV's
first-person view could never obtain.
Collaboration therefore requires no learned cross-agent representation at
all: the UAV supplies global spatial context, and the UGV performs
road-level planning and execution on top of it, each in its own role.

Our contributions are:
\begin{enumerate}
    \item \textbf{3D-SPF algorithm:} we upgrade SPF~\cite{spf2025} by adding a discrete
          height command (descend/hold/ascend) to the VLM's action interface,
          turning fixed-altitude planar point-fly into three-dimensional
          spatial search; relative to native UAV-only SPF, this lifts the
          UAV's success rate from 20.0\% to 60.0\%, a 3$\times$ gain, on our air-ground search task.
    \item \textbf{AGC-VLN system:} we couple 3D-SPF on the UAV with VLM
          road-path planning on the UGV through a \textbf{shared bird's-eye
          map} (the UAV renders the teammate's reported pose and the
          target's position), requiring no training or learned cross-agent
          representation. To the best of our knowledge, at the time of
          writing, no UAV+UGV collaboration VLN system with a positive collaboration gain
          has been reported in the open literature~\cite{carlaair_vla}, and AGC-VLN reaches a \textbf{77.0\%} joint
          success rate with a \textbf{+27.0\%} collaboration gain.
    \item \textbf{View-complementary collaboration mechanism:} we show that
          the UAV's global bird's-eye map fills the global spatial context
          missing from the UGV's first-person view, lifting the UGV's success
          rate from 12.0\% (UGV-only VLM) to 75.0\% (AGC-VLN), enabling
          road-level planning and closed-loop execution.
\end{enumerate}

\section{RELATED WORK}

\subsection{Foundation-Model Navigation Without Training}

The training-free paradigm replaces task-specific policies with frozen
foundation models orchestrated by deterministic glue.
SPF~\cite{spf2025} is the purest instance: a frozen VLM marks a waypoint
pixel and a depth label per frame, and fixed geometry turns the annotation
into a control command (no gradient ever flows).
Fly0~\cite{fly0} pushes the same semantic/geometric separation;
Uni-LaViRA~\cite{unilavira} drives four embodiments with a single
language-vision-action layer; FineCog-Nav~\cite{finecognav} splits reasoning
into seven inspectable modules.
The learned VLM-navigation line likewise replaces discrete action spaces
with continuous reasoning: NaVILA~\cite{navila} lifts VLM features into 3D,
MapGPT~\cite{mapgpt} grounds navigation in map-guided prompting,
GOAT~\cite{goat} generalizes goal specification to any object, and
StreamVLN~\cite{streamvln} streams history tokens for long horizons.
Large vision-language-action models (ABot-N1~\cite{abotn1},
LongNav-R1~\cite{longnavr1}, EvolveNav~\cite{evolvenav}) push toward one
general navigation policy, but stay trained and single-agent; none define
what two agents should \textit{say to each other}, the question this paper
answers.

\subsection{Cooperation Between Robots}

Simulation support for UAV+UGV teams arrived with
CARLA-Air~\cite{carlaair_infra}, which runs CARLA~\cite{carla} and
AirSim~\cite{airsim} in one Unreal Engine process with synchronized physics
ticks ($\Delta t = 0$\,ms).
The follow-up study~\cite{carlaair_vla} asked whether aerial VLA models can
cooperate, testing three coupling modes (none, UGV-to-UAV text hints (C1),
bidirectional velocity coupling (C2)) and finding that every mode failed to
improve (and often degraded) single-agent success across five models;
partner-state anchoring was isolated as the root deficit.
AirGroundBench~\cite{airgroundbench} supplies the evaluation material:
115 closed-loop episodes plus a four-level VQA taxonomy whose hardest level
probes cross-view reasoning.
On the method side, collaborative VLN exists only for \textit{homogeneous}
teams: two ground robots in CoNavBench~\cite{conavbench} and two UAVs at
different altitudes in AeroDuo~\cite{aeroduo}, whose high/low-viewpoint
division of labor echoes our view complementarity.
OmniVLN~\cite{omnivln} spans platforms but needs rotating LiDAR, and
JanusVLN~\cite{janusvln} decouples semantics from spatiality for
cross-platform navigation.
The aerial single-agent line has matured, from CityNav~\cite{citynav}
on real-world data to CLOSER-VLN~\cite{closervln} and
FSD-VLN~\cite{fsdvln} on long-horizon aerial VLN, but none couple an
aerial observer to a ground executor. Complementary benchmarks probe the
building blocks: SpatialUAV~\cite{spatialuav} (low-altitude collaboration),
an integrated planning framework~\cite{reliableagv} (air-ground field
deployment), and a survey~\cite{uavvlnsurvey} (aerial VLN).

\section{METHOD}

We formalize the task (Sec.~\ref{sec:problem}), describe the UAV's global
perception and map-rendering module (Sec.~\ref{sec:uav}) and the UGV's map
path-planning and closed-loop execution module (Sec.~\ref{sec:ugv}), and
give the two agents' parallel collaboration loop (Sec.~\ref{sec:comm}).

\begin{figure*}[t]
  \vspace*{8pt}
  \centering
  \includegraphics[width=0.99\textwidth]{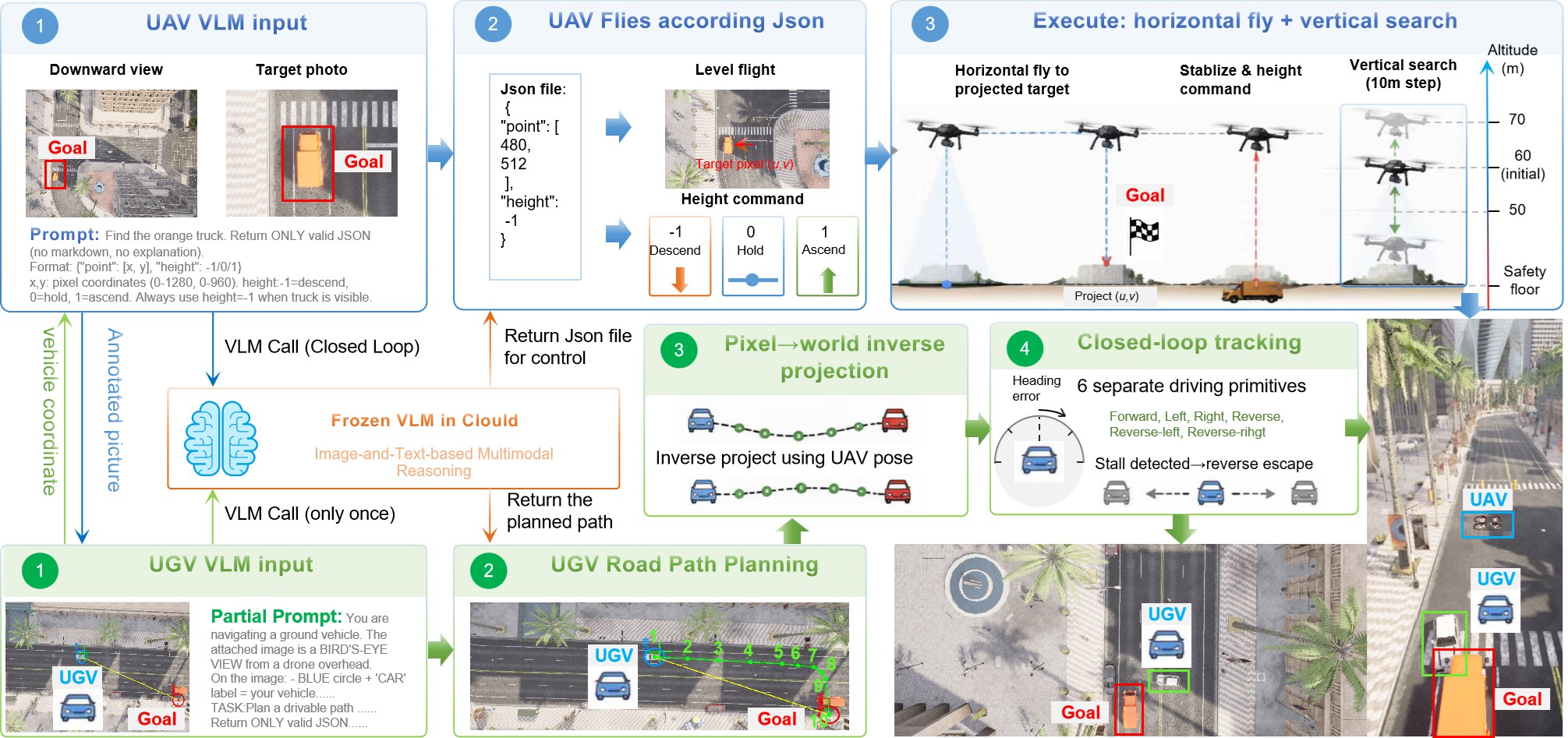}
  \caption{Architecture of AGC-VLN. On the UAV side, 3D-SPF
  queries a frozen VLM in closed loop to localize the
  target in the downward view, returning a JSON point
  and a height command (descend/hold/ascend) that drive horizontal flight and
  10\,m-step vertical search; it renders the UGV pose and the
  target's position as \texttt{CAR}/\texttt{GOAL} markers into a shared bird's-eye
  map. On the UGV side (right), a frozen VLM plans a road path on the
  map, which is inverse-projected to world coordinates and tracked by a
  closed-loop controller with six driving primitives and stuck recovery.}
  \label{fig:overview}
\end{figure*}

\subsection{Problem Formulation}
\label{sec:problem}

An episode hands a two-robot team a static target (a truck) specified by a prompt
description.
The team consists of a UAV hovering at altitude $h_{\mathrm{U}}$ whose RGB camera faces
straight down (intrinsics $\mathbf{K}_{\mathrm{U}}$), and a UGV whose RGB camera faces
forward (intrinsics $\mathbf{K}_{\mathrm{G}}$).
The episode succeeds if, before the time budget $T_{\max}$ expires, any team
member arrives within $\epsilon$ of the target without collision:
\begin{equation}
\text{success} \iff \min\big\{\|\mathbf{x}_{\mathrm{U}}-\mathbf{x}_{\mathrm{g}}\|_2,\ \|\mathbf{x}_{\mathrm{G}}-\mathbf{x}_{\mathrm{g}}\|_2\big\} \;\le\; \epsilon,
\label{eq:success}
\end{equation}
where $\mathbf{x}_{\mathrm{U}},\mathbf{x}_{\mathrm{G}}$ are the UAV and UGV positions and
$\mathbf{x}_{\mathrm{g}}$ the target position.
Both agents receive synchronized poses from the simulator; no map, no UAV depth
sensor, and no training data are assumed.
In this implementation $\epsilon = 5$\,m and $T_{\max} = 180$\,s.

\subsection{UAV Module: Global Perception and Map Rendering}
\label{sec:uav}

At each decision step (interval $\sim$3\,s, about 0.33\,Hz) the UAV performs
two functions: rendering the shared bird's-eye map for the UGV, and running
3D-SPF for itself.

\subsubsection{Bird's-Eye Annotation (Shared Map Rendering)}
The UAV reads the UGV's reported pose $\mathbf{x}_{\mathrm{G}}$ from the shared state
and projects $\mathbf{x}_{\mathrm{G}}$ onto the current downward image $I_{\mathrm{U}}^t$; the
target object's position is taken from the episode's goal coordinate
$\mathbf{x}_{\mathrm{g}}$ and projected onto $I_{\mathrm{U}}^t$.
The UAV then renders the annotations (Fig.~\ref{fig:overview}):
\begin{itemize}
    \item a blue \texttt{CAR} marker: the UGV's current position;
    \item a red \texttt{GOAL} marker: the target truck's position;
    \item a green cross: the UAV itself (image center);
    \item a yellow line: the CAR$\to$GOAL straight line (for reference only,
          not drivable);
    \item distance labels: UAV$\to$GOAL and CAR$\to$GOAL.
\end{itemize}
The fully annotated image constitutes a \textbf{shared bird's-eye map},
transmitted to the UGV through shared memory together with the UAV pose at
annotation time (used for deterministic pixel$\to$world inverse projection).

\subsubsection{3D-SPF (Three-Dimensional SPF)}
In parallel, the UAV runs 3D-SPF to fly toward the target itself: a frozen
VLM receives the downward view
$I_{\mathrm{U}}^t$, and returns structured JSON containing
the target truck's 2D pixel position $(u, v)$ in the downward view and a
discrete height command $h_c \in \{-1, 0, 1\}$ (descend/hold/ascend).
The pixel position is converted to a world coordinate $\mathbf{p}^{\mathrm{w}}$ via
flat-ground ray casting under a planar-ground assumption, parameterized by
the downward field of view $\theta$ and the altitude $h$:
\begin{equation}
\begin{aligned}
\mathbf{p}^{\mathrm{w}} &= \mathbf{c}_{\mathrm{U}} + \mathbf{R}(\psi)\,h\,\tan\!\tfrac{\theta}{2}
\begin{bmatrix} -2\big(\tfrac{v}{H}-\tfrac{1}{2}\big) \\[2pt] 2\big(\tfrac{u}{W}-\tfrac{1}{2}\big) \end{bmatrix},\\
\mathbf{R}(\psi) &= \begin{bmatrix}\cos\psi & -\sin\psi \\ \sin\psi & \cos\psi \end{bmatrix},
\end{aligned}
\label{eq:raycast}
\end{equation}
where $\psi$ is the yaw and $(W,H)$ the image size. The UAV flies
horizontally toward $\mathbf{p}^{\mathrm{w}}$ with proportional velocity
\begin{equation}
\mathbf{v} = \min\!\big(v_{\max},\; k_p\,\|\mathbf{p}^{\mathrm{w}}-\mathbf{c}_{\mathrm{U}}\|\big)\;
\frac{\mathbf{p}^{\mathrm{w}}-\mathbf{c}_{\mathrm{U}}}{\|\mathbf{p}^{\mathrm{w}}-\mathbf{c}_{\mathrm{U}}\|},
\label{eq:fly}
\end{equation}
then, once stabilized, descends/ascends 10\,m according to $h_c$ (with a
safety floor), then hovers awaiting the next decision.

\subsubsection{Coordinate Transformation}
The UAV and UGV live in different coordinate systems: AirSim uses NED ($z$
pointing down), CARLA uses a world frame ($z$ pointing up).
The two are converted through a constant offset $\mathbf{o}$ obtained
by one-time calibration:
\begin{equation}
\begin{aligned}
\mathbf{x}_{\mathrm{NED}} &= \mathbf{o} + \mathbf{S}\,\mathbf{x}_{\mathrm{CARLA}},
\qquad \mathbf{S} = \operatorname{diag}(1,1,-1),\\
\mathbf{o} &= \big(\mathrm{ap}_x-\mathrm{dl}_x,\; \mathrm{ap}_y-\mathrm{dl}_y,\; \mathrm{ap}_z+\mathrm{dl}_z\big)^{\top},
\end{aligned}
\label{eq:ned2carla}
\end{equation}
where $\mathrm{ap}$ and $\mathrm{dl}$ denote the drone position reported by
AirSim and CARLA, respectively, at calibration.
Pixel$\leftrightarrow$world conversion is done by flat-ground ray casting
(Eq.~\ref{eq:raycast} and its inverse), echoing cross-view geo-localization,
which aligns drone and overhead views~\cite{geo2,gengeo}.

\subsection{UGV Module: Path Planning and Closed-Loop Execution}
\label{sec:ugv}

Upon receiving the annotated bird's-eye map, the UGV executes three steps:

\subsubsection{Road Path Planning}
A frozen VLM combines the pixel coordinates of
CAR/GOAL in the map, and the image size to plan a path of 10 pixel waypoints
$\{\mathbf{w}_i\}_{i=1}^{10}$ along visible roads. The first point must equal
its own position and the last must equal the target position. The
intermediate points follow the road direction, turn at intersections, and
never cut through buildings or leave the road.
Before returning, the path is sanity-checked, discarding hallucinated paths
whose first/last points clearly deviate from CAR/GOAL.

\subsubsection{Pixel Path $\to$ World Path}
Using the UAV pose at annotation time, each pixel waypoint is
inverse-projected to NED coordinates, then converted through the offset
$\mathbf{o}$ to CARLA world coordinates, yielding the world path
$\{\mathbf{w}_i^{\mathrm{w}}\}$.

\subsubsection{Closed-Loop Path Tracking}
The UGV tracks the world path point by point with a closed-loop controller:
at each tick it computes the wrapped heading error to the current waypoint
and a distance-scaled throttle,
\begin{equation}
e_{\psi} = \operatorname{wrap}_{\pi}\!\big(\psi_{\mathrm{wp}}-\psi\big),
\qquad
\tau = \operatorname{clip}\!\big(\tfrac{d}{10},\;0.3,\;1\big),
\label{eq:follow}
\end{equation}
where $\psi$ is the UGV heading, $\psi_{\mathrm{wp}}$ the heading toward the
waypoint, and $d$ the remaining distance. The error $e_{\psi}$ is discretized
into six driving primitives (forward, left, right, reverse, reverse-left,
reverse-right) by angle thresholds. When the vehicle stalls beyond a
threshold (throttle applied but near-zero displacement), a reverse escape is
triggered.
After reaching each waypoint, it checks whether it has entered the
$\epsilon$ range of the target.

\begin{algorithm}[t]
\caption{AGC-VLN: Training-Free Air-Ground Collaborative VLN (Bird's-Eye Map Sharing)}
\label{alg:baseline}
\begin{algorithmic}[1]
\REQUIRE Goal position $\mathbf{x}_{\mathrm{g}}$, budget $T_{\max}$
\ENSURE Episode outcome $\in$ \{success, failure\}
\STATE Start UAV and UGV threads in parallel, sharing state $\mathcal{S}$; clear both success flags
\WHILE{$t < T_{\max}$ and not both agents succeeded}
    \STATE \textbf{UAV thread:} read $\mathcal{S}.\text{ugv\_pos}$; write its own pose
    \STATE \quad $(u, v, h_c) \gets \text{VLM.Locate}(I_{\mathrm{U}}^t)$ \hfill // target pixel $(u,v)$; height $h_c \in \{-1,0,1\}$
    \STATE \quad $M^t \gets \text{Annotate}(I_{\mathrm{U}}^t, \mathbf{x}_{\mathrm{G}}, \mathbf{x}_{\mathrm{g}})$ \hfill // CAR + GOAL markers; GOAL at the target position $\mathbf{x}_{\mathrm{g}}$
    \STATE \quad $\mathcal{S}.\text{map} \gets (M^t, \text{pose}_t)$ \hfill // shared bird's-eye map + pose at annotation time
    \STATE \quad fly toward $\mathbf{p}^{\mathrm{w}} = \text{Project}(u,v)$; adjust altitude by $h_c$ \hfill // 3D-SPF flight
    \STATE \quad if UAV within $\epsilon$ of target: mark UAV success (keep annotating)
    \STATE \textbf{UGV thread:} read $\mathcal{S}.\text{map}$
    \STATE \quad $\{\mathbf{w}_i\}_{i=1}^{10} \gets \text{VLM.PlanPath}(M^t)$ \hfill // road path planning
    \STATE \quad $\{\mathbf{w}_i^{\mathrm{w}}\} \gets \text{InverseProject}(\{\mathbf{w}_i\}, \text{pose}_t)$
    \STATE \quad $\text{FollowPath}(\{\mathbf{w}_i^{\mathrm{w}}\})$ \hfill // closed-loop tracking + stuck recovery
    \STATE \quad if UGV within $\epsilon$ of target: mark UGV success (keep reporting pose)
\ENDWHILE
\RETURN success if either agent marked success, else failure
\end{algorithmic}
\end{algorithm}

\subsection{Collaboration Mechanism}
\label{sec:comm}

Algorithm~\ref{alg:baseline} summarizes the complete parallel loop. The two
agents collaborate asynchronously through shared memory, exchanging two kinds
of information:
\begin{itemize}
    \item \textbf{UGV$\to$UAV (pose)}: the UGV continuously reports its
          CARLA world coordinate, for the UAV to render the \texttt{CAR}
          marker;
    \item \textbf{UAV$\to$UGV (shared bird's-eye map)}: the UAV continuously
          outputs the annotated bird's-eye image and pose at annotation
          time, for the UGV to plan a path.
\end{itemize}
The key to the collaboration is \textbf{view complementarity}. The UAV has a
global bird's-eye view but cannot drive along roads, while the UGV can drive
along roads but has only a first-person local view. The annotated map
delivers the UAV's global spatial context (relative positions of teammate
and target, road topology) in an image form the UGV's VLM can consume
directly, thereby filling the UGV's blind spot.

\section{EXPERIMENTS}

We design experiments to answer three research questions:
\begin{enumerate}
    \item \textbf{RQ1 (Feasibility):} Can training-free single-agent methods
          be composed into a working air-ground collaborative VLN system,
          where trained VLA models failed~\cite{carlaair_vla}?
    \item \textbf{RQ2 (Collaboration gain):} Does the UAV-UGV team outperform
          each agent operating alone, and by how much?
    \item \textbf{RQ3 (Failures):} Where does the baseline still fail, and
          what do the failures imply for future learned components?
\end{enumerate}

\subsection{Experimental Setup}

\textbf{Testbed:} simulation experiments run inside
CARLA-Air~\cite{carlaair_infra}, whose single-process design guarantees
that UAV and UGV observations sample the same physics tick.
The quadrotor carries a downward 1080p RGB camera (FOV 108$^\circ$),
hovering at 60\,m; the UGV carries a forward camera; both read poses from
the synchronized pose stream.

\textbf{Episodes:} 100 closed-loop episodes across 50 spawn points (2 runs each) in the Town10HD scene; a Mini Cooper (UGV) is spawned at the start and an HGV truck at the
goal.

\textbf{Metrics:} we report the UAV success rate ($\text{SR}_{\text{UAV}}$,
the UAV arrives within $\epsilon$ of the target), the UGV success rate
($\text{SR}_{\text{UGV}}$), and the joint success rate
($\text{SR}_{\text{joint}}$, either member arrives), with $\epsilon = 5$\,m.
We also report Success weighted by Path Length (SPL), Navigation Error
(NE), and the \textbf{collaboration gain},
$\text{CG} = \text{SR}_{\text{joint}} - \min(\text{SR}_{\text{UAV}},
\text{SR}_{\text{UGV}})$, which measures how much the team exceeds the
weaker single agent.

\textbf{Configuration:} \texttt{gemini-3.7-flash}~\cite{gemini} (temperature 0.1) serves as the
frozen VLM on both platforms; the UAV/UGV decision interval is 3\,s
($\sim$0.33\,Hz); the time budget is 180\,s.

\subsection{Baselines}

\textbf{(1) Published single-agent aerial VLN methods:} OpenFly~\cite{openfly},
FineCog-Nav~\cite{finecognav}, 3DG-VLN~\cite{seereach}, and Travel
UAV~\cite{traveluav}, reproduced under the same single-agent evaluation
protocol for comparison.

\textbf{(2) Single-agent (collaboration lower bound):}
UAV-only SPF, UAV-only 3D-SPF, and UGV-only VLM, each solving the whole
episode alone.

\subsection{Main Results (RQ1, RQ2)}
\label{sec:main_results}

\begin{table*}[t]
\vspace*{8pt}
\caption{Main results of air-ground collaborative VLN (100 episodes). See the note below the table for column definitions.}
\label{tab:main_results}
\begin{center}
\small
\setlength{\tabcolsep}{1.8pt}
\begin{tabular}{lcccccccccc}
\toprule
\rowcolor[HTML]{D9E2F3}\textbf{Method} & \textbf{SR$_{\text{UGV}}$} $\uparrow$ & \textbf{SR$_{\text{UAV}}$} $\uparrow$ & \textbf{SR$_{\text{joint}}$} $\uparrow$ & \textbf{SPL} $\uparrow$ & \textbf{NE (m)} $\downarrow$ & \textbf{CG} $\uparrow$ & \textbf{Time (s)} $\downarrow$ & \textbf{VLM calls} $\downarrow$ & \textbf{Path$_{\text{UAV}}$ (m)} $\downarrow$ & \textbf{Path$_{\text{UGV}}$ (m)} $\downarrow$ \\
\midrule
\rowcolor[HTML]{F2F2F2}\multicolumn{11}{c}{\textit{Published single-agent VLN baselines}} \\
\midrule

OpenFly~\cite{openfly}  & --- & 0.0\% & 0.0\% & 0.0\% & 78.4$\pm$18.7 & 0.0\% & --- & 34.1$\pm$7.3 & 56.4$\pm$16.0 & --- \\
FineCog-Nav~\cite{finecognav}    & --- & 13.0\% & 13.0\% & 10.0\% & 28.4$\pm$12.9 & 0.0\% & 177.2$\pm$35.4 & 58.0$\pm$12.4 & 19.9$\pm$10.4 & --- \\
3DG-VLN~\cite{seereach}  & --- & 33.0\% & 33.0\% & 26.0\% & 49.6$\pm$32.1 & 0.0\% & 142.1$\pm$60.4 & 11.7$\pm$5.2 & 49.2$\pm$21.7 & --- \\
Travel UAV~\cite{traveluav}    & --- & 53.0\% & 53.0\% & 45.0\% & 30.3$\pm$24.3 & 0.0\% & 115.8$\pm$63.4 & 21.0$\pm$11.5 & 42.24$\pm$27.70 & --- \\
\midrule
\rowcolor[HTML]{F2F2F2}\multicolumn{11}{c}{\textit{Single-agent ablations (our components)}} \\
\midrule
UGV-only VLM      & 12.0\% & --- & 12.0\% & 11.6\% & 115.2$\pm$65.6 & 0.0\% & 28.2$\pm$3.0 & 12.1$\pm$4.0 & --- & 49.4$\pm$4.0 \\
UAV-only SPF~\cite{spf2025}      & --- & 20.0\% & 20.0\% & 18.0\% & 155.5$\pm$94.9 & 0.0\% & 58.5$\pm$9.6 & 12.9$\pm$6.5 & 119.1$\pm$74.1 & --- \\
UAV-only 3D-SPF   & --- & 60.0\% & 60.0\% & 46.0\% & 30.0$\pm$40.9 & 0.0\% & 104.0$\pm$8.0 & 13.0$\pm$7.1 & 84.3$\pm$36.7 & --- \\
\midrule
\rowcolor[HTML]{F2F2F2}\multicolumn{11}{c}{\textit{Ours: Air-Ground Collaborative VLN}} \\
\midrule
\rowcolor[HTML]{FFF2CC}\textbf{AGC-VLN (ours)} & \textbf{75.0\%} & \textbf{50.0\%} & \textbf{77.0\%} & \textbf{62.0\%} & \textbf{10.9$\pm$16.2} & \textbf{+27.0\%} & \textbf{82.2$\pm$27.0} & \textbf{12.1$\pm$2.9} & \textbf{65.1$\pm$22.2} & \textbf{43.5$\pm$14.0} \\
\bottomrule
\end{tabular}
\par\vspace{6pt}
{\footnotesize
\textbf{Note:} $\uparrow$/$\downarrow$ = higher/lower is better. \textbf{SR$_{\text{UAV}}$/SR$_{\text{UGV}}$/SR$_{\text{joint}}$}: UAV/UGV/joint success rate (goal within 5\,m). \textbf{SPL}: success weighted by path length; \textbf{NE}: navigation error; \textbf{CG}: collaboration gain $=\text{SR}_{\text{joint}}-\min(\text{SR}_{\text{UAV}},\text{SR}_{\text{UGV}})$; \textbf{Time}: mean arrival time over successful episodes; \textbf{VLM calls}: inference count; \textbf{Path$_{\text{UAV}}$/Path$_{\text{UGV}}$}: distance traveled. SR/SPL/CG are percentages; NE/VLM calls/path are mean$\pm$std over all episodes; Time is mean$\pm$std over successful episodes; ``---'' = not applicable/reported.}
\end{center}
\end{table*}

Table~\ref{tab:main_results} reports the main results. We measure the collaboration gain as the amount by which the joint success
rate exceeds the weaker individual agent:
\begin{equation}
\mathrm{CG} = \mathrm{SR}_{\text{joint}} - \min\!\big(\mathrm{SR}_{\text{UAV}},\,\mathrm{SR}_{\text{UGV}}\big).
\label{eq:cg}
\end{equation}
Two findings follow: (1)~\textbf{positive collaboration gain}. AGC-VLN
attains $\text{SR}_{\text{joint}} = 77.0\%$ ($\text{SR}_{\text{UGV}} = 75.0\%$,
$\text{SR}_{\text{UAV}} = 50.0\%$), a collaboration gain of
$\text{CG} = +27.0\%$ over the weaker agent, together with an SPL of $62.0\%$
and an NE of $10.9$\,m; the joint rate thereby exceeds the strongest published
single-agent baseline (Travel UAV, $53.0\%$) by $24.0$ points, unlike the VLA
baselines whose coupling degraded performance; (2)~\textbf{success means
either member arrives}. The task is judged successful if either agent reaches
the target, so $\text{SR}_{\text{joint}}$ is the \emph{union} of the per-agent
rates.

\subsection{Ablation: VLM Backbone}

Following SPF's cross-VLM study, we swap the frozen VLM on both platforms
across four backbones (\texttt{gpt-5.6-luna}, \texttt{gemini-2.5-flash},
\texttt{gemini-3.7-flash}, and \texttt{qwen-vl-max}~\cite{gemini}) and
report success rates and costs (Table~\ref{tab:backbone}).
\texttt{gemini-3.7-flash} attains the highest joint success rate (77.0\%) as
the most balanced backbone, keeping both agents strong
($\text{SR}_{\text{UGV}} = 75.0\%$, $\text{SR}_{\text{UAV}} = 50.0\%$); the
alternatives leave the UAV weaker ($\text{SR}_{\text{UAV}} = 7.0\%$--$47.0\%$),
so their joint rates fall short ($67.0\%$--$73.0\%$).

\begin{table}[t]
\caption{VLM Backbone Ablation.}
\label{tab:backbone}
\begin{center}
\footnotesize
\setlength{\tabcolsep}{1.7pt}
\begin{tabular}{lccccc}
\toprule
\rowcolor[HTML]{D9E2F3}\textbf{VLM} & \textbf{SR$_{\text{UGV}}$} $\uparrow$ & \textbf{SR$_{\text{UAV}}$} $\uparrow$ & \textbf{SR$_{\text{joint}}$} $\uparrow$ & \textbf{CG} $\uparrow$ & \textbf{Time (s)} $\downarrow$ \\
\midrule
qwen-vl-max & 73.0\% & 47.0\% & 73.0\% & +26.0\% & 75.23$\pm$68.80 \\
gpt-5.6-luna       & 67.0\% & 13.0\% & 67.0\% & +54.0\% & 102.97$\pm$63.10 \\
gemini-2.5-flash   & 67.0\% & 7.0\% & 73.0\% & +66.0\% & 91.91$\pm$63.36 \\
\rowcolor[HTML]{FFF2CC}\textbf{gemini-3.7-flash} & \textbf{75.0\%} & \textbf{50.0\%} & \textbf{77.0\%} & \textbf{+27.0\%} & \textbf{82.2$\pm$27.0} \\
\bottomrule
\end{tabular}
\end{center}
\end{table}

\subsection{Ablation: UAV Altitude}

To test 3D-SPF's sensitivity to the UAV's initial altitude, we vary it across
30/60/90/120\,m while keeping the UGV module and the height command
(descend/hold/ascend) unchanged (Table~\ref{tab:height_ablation}).
3D-SPF is most effective at 60\,m, where both agents contribute: the UAV's
success rate peaks at 50.0\% and the UGV's road-following reaches 75.0\%,
yielding the best joint rate of 77.0\%. At 30\,m the narrowed field of view
drops the UAV to 20.0\% and the UGV to 14.0\%, collapsing the joint rate to
28.0\%; at 90/120\,m the truck shrinks to a handful of pixels and UAV
localization fails entirely (0.0\%), so the high joint rate (77.0\%/72.0\%)
is delivered by the UGV alone, with no aerial contribution.

\begin{table}[t]
\caption{3D-SPF Altitude Ablation}
\label{tab:height_ablation}
\begin{center}
\small
\setlength{\tabcolsep}{1.5pt}
\begin{tabular}{lccccc}
\toprule
\rowcolor[HTML]{D9E2F3}\textbf{Initial altitude} & \textbf{SR$_{\text{UGV}}$} $\uparrow$ & \textbf{SR$_{\text{UAV}}$} $\uparrow$ & \textbf{SR$_{\text{joint}}$} $\uparrow$ & \textbf{CG} $\uparrow$ & \textbf{Time (s)} $\downarrow$ \\
\midrule
30\,m  & 14.0\% & 20.0\% & 28.0\% & +14.0\%  & 141.3$\pm$29.7 \\
\rowcolor[HTML]{FFF2CC}\textbf{60\,m (Ours)} & \textbf{75.0\%} & \textbf{50.0\%} & \textbf{77.0\%} & \textbf{+27.0\%} & \textbf{82.2$\pm$27.0} \\
90\,m  & 77.0\% & 0.0\%  & 77.0\% & +77.0\% & 66.1$\pm$29.5 \\
120\,m & 72.0\% & 0.0\%  & 72.0\% & +72.0\% & 70.6$\pm$35.0 \\
\bottomrule
\vspace{-0.5cm}
\end{tabular}
\end{center}
\end{table}

\subsection{Ablation: Map Annotation Richness}

\begin{figure*}[t]
  \centering
  \includegraphics[width=0.99\textwidth]{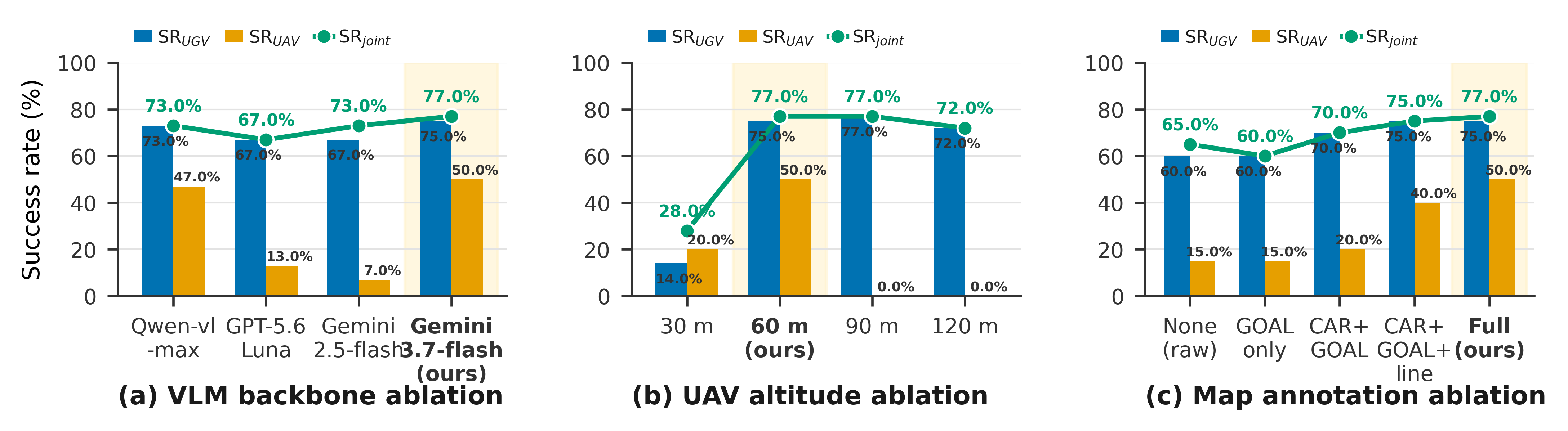}
  \caption{Ablation across the three design axes. (a)~VLM backbone (Qwen-vl-max,
  GPT-5.6 Luna, Gemini 2.5-flash, and Gemini 3.7-flash). (b)~UAV initial
  altitude (30/60/90/120\,m). (c)~Map annotation richness (raw, goal-only,
  CAR+GOAL, CAR+GOAL+line, and full). Each panel reports
  SR$_{\text{joint}}$ (green line) together with the per-agent success rates
  SR$_{\text{UGV}}$ and SR$_{\text{UAV}}$ (bars), with percentage value labels
  and the adopted configuration (``ours'') highlighted.}
  \label{fig:ablation}
\end{figure*}

To test the necessity of the shared map's annotation content, we degrade the
bird's-eye annotation stepwise through five levels of richness: full
annotation (CAR+GOAL+distance labels+reference line), CAR+GOAL+reference
line, CAR+GOAL, GOAL only, and finally a completely unannotated raw
bird's-eye image (Table~\ref{tab:annot_ablation}).
Adding the CAR marker lifts the UGV's success rate from 60.0\% (GOAL only)
to 70.0\% (CAR+GOAL), the reference line further to 75.0\%, and full
annotation (with distance labels) reaches 77.0\% joint success. Notably,
GOAL-only (60.0\%) underperforms the unannotated raw image (65.0\%), indicating
that a lone GOAL marker without the teammate's CAR anchor misleads the UGV's
path planner.
Figure~\ref{fig:ablation} summarizes the three ablation
axes.

\begin{table}[t]
\vspace*{8pt}
\caption{Map Annotation Richness Ablation (bird's-eye map sharing)}
\label{tab:annot_ablation}
\begin{center}
\small
\setlength{\tabcolsep}{1pt}
\begin{tabular}{lccccc}
\toprule
\rowcolor[HTML]{D9E2F3}\textbf{Map annotation} & \textbf{SR$_{\text{UGV}}$} $\uparrow$ & \textbf{SR$_{\text{UAV}}$} $\uparrow$ & \textbf{SR$_{\text{joint}}$} $\uparrow$ & \textbf{CG} $\uparrow$ & \textbf{Time (s)} $\downarrow$ \\
\midrule
None (raw)               & 60.0\% & 15.0\% & 65.0\% & +50.0\% & 118.8$\pm$53.4 \\
GOAL only                & 60.0\% & 15.0\% & 60.0\% & +45.0\% & 116.0$\pm$55.4 \\
CAR+GOAL               & 70.0\% & 20.0\% & 70.0\% & +50.0\% & 106.2$\pm$51.8 \\
CAR+GOAL+line        & 75.0\% & 40.0\% & 75.0\% & +35.0\% & 95.6$\pm$50.6 \\
\rowcolor[HTML]{FFF2CC}\textbf{Full (ours)}     & \textbf{75.0\%} & \textbf{50.0\%} & \textbf{77.0\%} & \textbf{+27.0\%} & \textbf{82.2$\pm$27.0} \\
\bottomrule
\end{tabular}
\end{center}
\end{table}

\subsection{Failure Source Analysis (RQ3)}
\label{sec:qualitative}

\begin{figure}[t]
  \centering
  \includegraphics[width=\columnwidth]{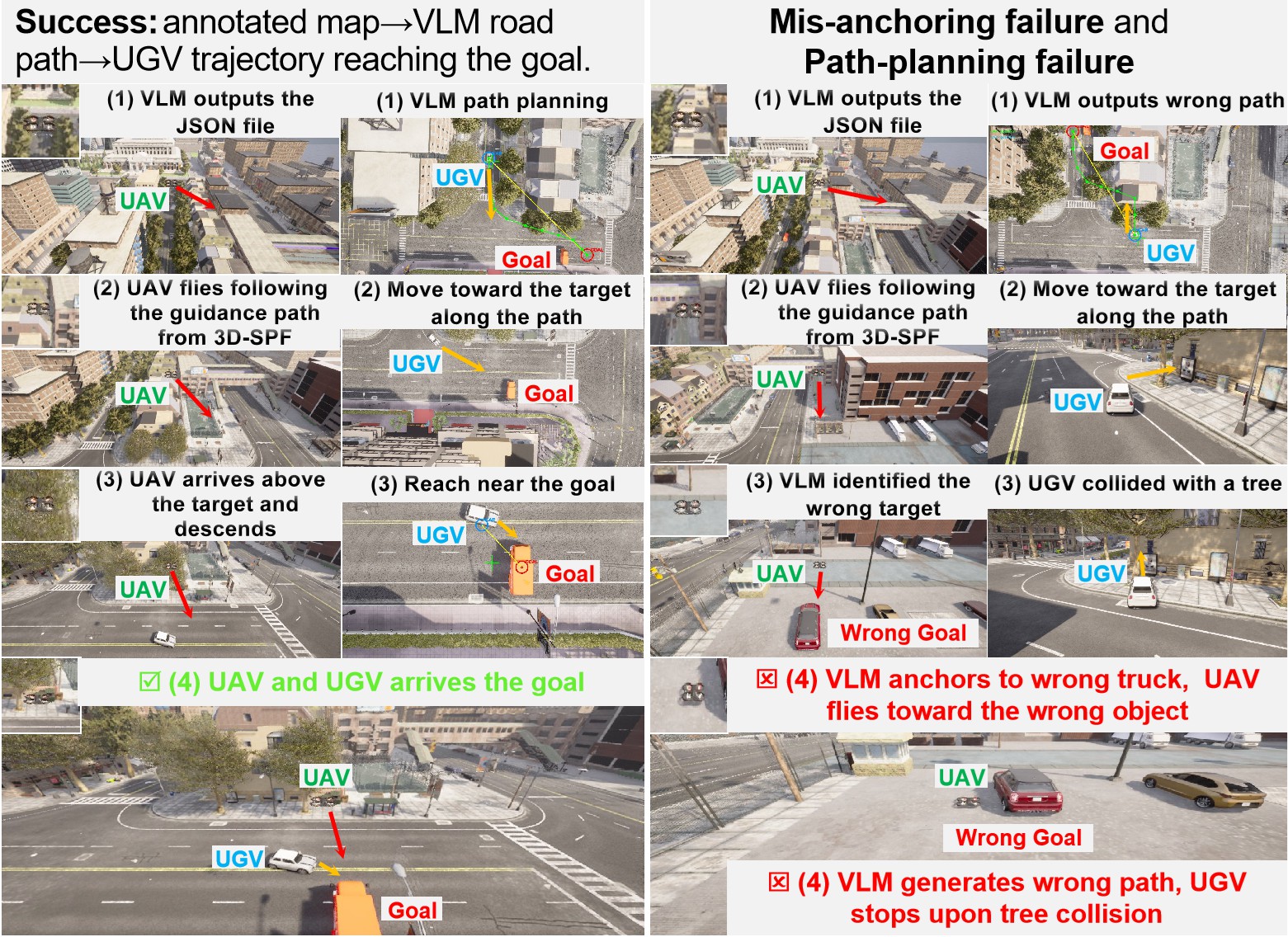}
  \caption{Representative episodes. Success: the annotated map drives a VLM
  road path that the UGV tracks to the goal, with the UAV and UGV both
  arriving. Mis-anchoring failure: the UAV's VLM anchors onto a similar but
  wrong truck and flies toward the wrong object. Path-planning failure: the
  UGV's VLM plans a path that collides with a tree, stalling the UGV.}
  \label{fig:qualitative}
\end{figure}

Figure~\ref{fig:qualitative} shows representative episodes. We attribute each
failed episode to the \textit{first} pipeline stage that deviated from ground
truth (Table~\ref{tab:failures}): \textbf{global localization} (the UAV's VLM
marks a wrong target region), \textbf{path planning} (an undrivable path), or
\textbf{local execution} (failing to close the final meters).

\begin{table}[t]
\vspace*{8pt}
\caption{Failure Attribution by Pipeline Stage}
\label{tab:failures}
\begin{center}
\small
\setlength{\tabcolsep}{4pt}
\begin{tabular}{p{1in}cp{1.75in}}
\toprule
\rowcolor[HTML]{D9E2F3}\textbf{Stage} & \textbf{Share} & \textbf{Typical case} \\
\midrule
\rowcolor[HTML]{F2F2F2}\textbf{Global localization} & 17\% & \\
\quad{mis-anchoring}   & 4\%  & locks onto a wrong target \\
\quad{unstable tracking} & 13\% & points jump across frames \\
\addlinespace
\rowcolor[HTML]{F2F2F2}\textbf{Path planning}   & 61\% & undrivable path; vehicle crashes and stalls \\
\rowcolor[HTML]{F2F2F2}\textbf{Local execution} & 22\% & final-approach dead end or timeout \\
\midrule
\rowcolor[HTML]{FFF2CC}\textbf{Total}  & 100\% & 23 jointly-failed episodes \\
\bottomrule
\end{tabular}
\end{center}
\end{table}

Measured across the $23$ jointly-failed episodes, path planning is the
dominant stage ($14/23=61\%$; Table~\ref{tab:failures}): the UGV's VLM emits
a path that collapses onto a building, tree, or the map border, and the
vehicle crashes and stalls $16$--$73$~m from the goal. Local execution
($5/23=22\%$) closes within $15$~m but hits the $180$~s limit. Global
localization is the residual UAV-side failure ($4/23=17\%$), split between
unstable tracking ($3/23=13\%$), whose points jump across frames, and
mis-anchoring ($1/23=4\%$), the reference-binding failure we had
hypothesized, in which the UAV locks onto a similar but wrong truck and flies
toward it. Map rendering/projection contributed no failures and is omitted.

Per-agent, the UAV fails from global localization ($76\%$) and the UGV from
path planning ($60\%$); the UGV rescues $27$ of the UAV's $50$ failures versus
$2$ the other way. Cross-view target re-identification is the
highest-leverage learned component~\cite{airgroundbench}.

\section{REAL-ROBOT EXPERIMENT CASE}

Beyond the simulation, we deploy the same training-free pipeline on real
hardware to verify its feasibility once decoupled from the simulator's
synchronized pose stream.
A UAV (downward RGB camera) and a UGV
(forward camera) form the team. The UAV runs 3D-SPF, and the
UGV runs road-level path planning on the shared bird's-eye map.
The frozen VLM remains \texttt{gemini-3.7-flash}, and the decision cadence
matches the simulation.
In the deployment the UAV pose is provided by
LiDAR odometry, while the stages
(annotation, projection, closed-loop execution) are identical to the
simulation.

\begin{figure}[t]
  \centering
  \includegraphics[width=\columnwidth]{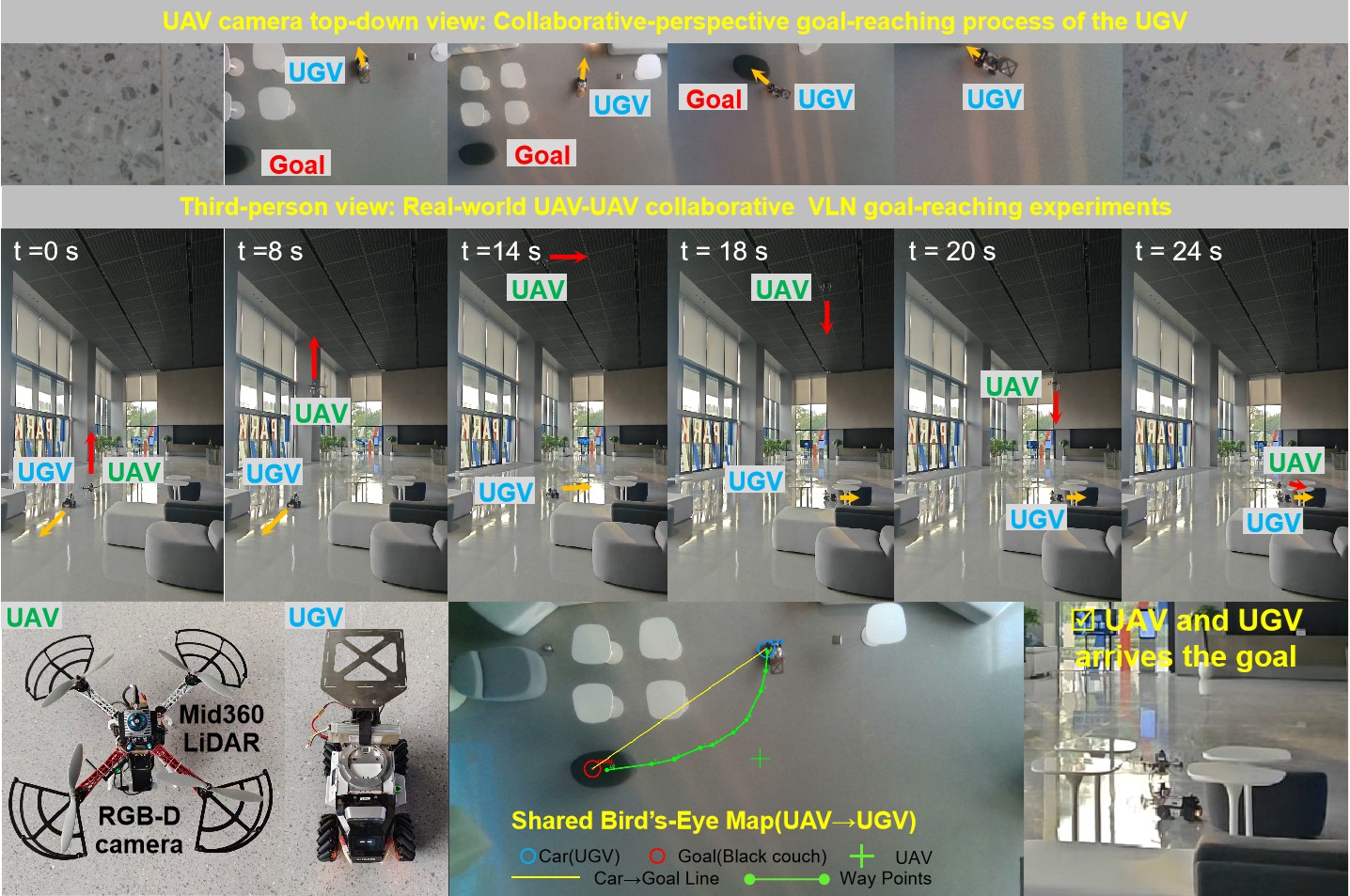}
  \caption{Real-robot case: the same training-free pipeline (3D-SPF + shared
  bird's-eye map road planning) running on a physical UAV+UGV team. Six time steps in top-down and third-person views show the quadrotor (Mid360 LiDAR + RealSense D435i) annotating the bird's-eye map and flying 3D-SPF toward the target, while the omnidirectional robot follows its VLM-planned road path.}
  \label{fig:realrobot}
\end{figure}

The physical deployment confirms that the training-free pipeline drives both
agents to the target, as shown in Figure~\ref{fig:realrobot}, which depicts
the real-world air-ground collaborative VLN hardware and scene. The UAV runs
Linux and carries a Mid360 LiDAR and a RealSense D435i, while the UGV carries
a forward camera. The scene spans six time steps in both top-down and
third-person
views: the UAV first climbs to altitude, calls the VLM to annotate the
bird's-eye view, and sends it to the UGV, which calls the VLM to plan a road
path (the green point set); the UAV then approaches the target with the
proposed 3D-SPF, while the UGV automatically steers along the planned path and
moves toward the target.

\section{LESSONS AND LIMITATIONS}

\subsection{Lessons for Air-Ground Collaboration}

\textbf{Lesson 1: put the interface at the global view.}
The UAV's bird's-eye view naturally carries the global relative positions of
teammate and target, spatial relationships that mean the same thing in both
agents' frames, so choosing it as the communication interface (rather than
text, features, or control signals) is what turns collaboration from hurting
into helping relative to~\cite{carlaair_vla}.

\textbf{Lesson 2: view complementarity, not message richness.}
The gain comes not from transmitting richer information but from supplying
the viewpoint the receiver lacks: the global spatial context invisible in the
UGV's first-person view, which lets the UGV's VLM turn the UAV's ``can see''
into the UGV's ``can drive''.

\textbf{Lesson 3: determinism confines error propagation.}
Because projection and inverse projection contain no learned parts, a VLM
mistake on one platform cannot propagate to the other, structurally the
opposite of the C2 velocity coupling that amplified errors
in~\cite{carlaair_vla}.

\subsection{Limitations}

The baseline adopts three deliberate simplifications. The shared map marks the
target with the episode-provided goal position, which keeps the collaboration
interface deterministic, while the UAV's own flight (3D-SPF) still inherits the
VLM's mis-anchoring risk. The 3\,s VLM latency confines the agents to quasi-static scenes, leaving
fast-moving targets out of reach. And CARLA-Air's synchronized poses stand in
for the GPS/SLAM a real field deployment would need. Each simplification is
intentional: a minimal baseline keeps the remaining gaps measurable.

\section{CONCLUSION}

We presented \textbf{AGC-VLN} (Air-Ground Collaborative VLN), the first
training-free baseline for air-ground collaborative VLN, which couples 3D-SPF
on the UAV with VLM road-path planning on the UGV through a \textbf{shared
bird's-eye map}. On 100 closed-loop episodes in Town10HD it achieves a clear
positive collaboration gain, in contrast to trained VLA models whose coupling
degraded single-agent performance. Residual failures concentrate on global
localization, chiefly mis-anchoring onto a wrong truck (reference
binding), which delineates a concrete agenda: robust cross-view target
re-identification, learned verification, and latency-robust coordination. We
release the full system as a reproducible starting point for the field.


\bibliographystyle{IEEEtran}
\bibliography{references}

\end{document}